\documentclass[twoside]{article}

\usepackage[accepted]{aistats2021}
\usepackage{graphicx}
\usepackage{amssymb}
\usepackage{amsthm}
\usepackage{float}
\usepackage{dblfloatfix} 
\usepackage{natbib}
\usepackage{ dsfont }
\usepackage{algorithm}
\usepackage[noend]{algpseudocode}
\usepackage{url}
\newtheorem{proposition}{Proposition}

\begin{document}

%

%

\twocolumn[

\aistatstitle{ABC-Di: Approximate Bayesian Computation for Discrete Data}

\aistatsauthor{ Ilze Amanda Auzina \And Jakub M. Tomczak}

\aistatsaddress{ Vrije Universiteit Amsterdam \\ \texttt{ilze.amanda.auzina@gmail.com} \And Vrije Universiteit Amsterdam \\ \texttt{jmk.tomczak@gmail.com} } ]

\begin{abstract}
Many real-life problems are represented as a black-box, i.e., the internal workings are inaccessible or a closed-form mathematical expression of the likelihood function cannot be defined. For continuous random variables likelihood-free inference problems can be solved by a group of methods under the name of Approximate Bayesian Computation (ABC). However, a similar approach for discrete random variables is yet to be formulated. Here, we aim to fill this research gap. We propose to use a population-based MCMC ABC framework. Further, we present a valid Markov kernel, and propose a new kernel that is inspired by Differential Evolution. We assess the proposed approach on a problem with the known likelihood function, namely, discovering the underlying diseases based on a QMR-DT Network, and three likelihood-free inference problems: (i) the QMR-DT Network with the unknown likelihood function, (ii) learning binary neural network, and (iii) Neural Architecture Search. The obtained results indicate the high potential of the proposed framework and the superiority of the new Markov kernel.
\end{abstract}

\section{Introduction}

In various scientific domains an accurate simulation model can be designed, yet formulating the corresponding likelihood function remains a challenge. In other words, there is a simulator of a process available that when provided an input, returns an output, but the inner workings of the process are not analytically available \citep{audet2017derivative, beaumont2002approximate, cranmer2020frontier, lintusaari2017fundamentals, toni2009approximate}. Thus far, the existing tools for solving such problems are typically limited to continuous random variables. Consequently, many discrete problems are reparameterized to continuous ones via, for example, the Gumbel-Softmax trick \citep{jang2016categorical} rather than solved directly. In this paper, we aim at providing a solution to this problem by translating the existing likelihood-free inference methods to discrete space applications. 

A group of methods used to solve likelihood-free inference problems for continuous data is known under the name of Approximate Bayesian Computation (ABC) \citep{beaumont2002approximate}. The main idea behind any ABC method is to model the posterior distribution by approximating the likelihood as a fraction of accepted simulated data points from the simulator model, by the use of a distance measure $\delta$ and a tolerance value $\epsilon$. The first approach, known as the ABC-rejection scheme, was successfully applied in biology \citep{pritchard1999population, tavare1997inferring} and since then many alternative versions of the algorithm have been introduced, with the three main groups represented by Markov Chain Monte Carlo ABC \citep{marjoram2003markov}, Sequential Monte Carlo (SMC) ABC \citep{beaumont2009adaptive}, and neural-network-based ABC \citep{papamakarios2019neural, papamakarios2019sequential}. Here, we focus on the MCMC-ABC version \citep{andrieu2009pseudo} as it can be more readily implemented and the computational costs are lower \citep{jasra2007population}. As a result, the efficiency of the newly proposed method will depend on two parts, namely, (i) a proposal distribution for the MCMC algorithm, and (ii) a selection of hyperparameter values for the ABC algorithm. 

Here, we are interested in likelihood-free inference in discrete spaces where there is no ''natural'' notion of search direction and scale.
Therefore, our main focus in on an efficient Markov kernel deisgn in a population-based MCMC framework. The presented solution is inspired by differential evolution (DE) \citep{storn1997differential} that has been shown to be an effective optimization technique for many likelihood-free (or black-box) problems \citep{vesterstrom2004comparative, mauvcec2018improved}. We propose to define a probabilistic DE kernel for discrete random variables that allows us to traverse the search space without specifying any external parameters. We evaluate our approach on four test-beds: (i) we verify our proposal on a benchmark problem of QMR-DT Network presented by \cite{strens2003evolutionary}; (ii) we modify the first problem and formulate it as a likelihood-free inference problem; (iii) we assess the applicability of our method for high-dimensional data, namely, training binary neural networks on MNIST data; (iv) we apply the proposed approach to Neural Architecture Search (NAS) using the benchmark dataset proposed by \cite{ying2019bench}.

The contribution of the present paper is as follows. First, we introduce an alternative version of the MCMC-ABC algorithm, namely, a population-based MCMC-ABC method, that is applicable to likelihood-free inference tasks with discrete random variables. Second, we propose a novel Markov kernel for likelihood-based inference methods in a discrete state space. Third, we present the utility of the proposed approach on three binary problems.

\section{Likelihood-free inference and ABC}
Let $x \in \mathcal{X}$ be a vector of parameters or decision variables, where $\mathcal{X} = \mathbb{R}^{D}$ or $\mathcal{X} = \{0, 1\}^{D}$, and $y \in \mathbb{R}^{M}$ is a vector of observable variables. Typically, for a given collection of observations of $y$, $y_{data}=\{y_{n}\}_{n=1}^{N}$, we are interested in solving the following optimization problem:\footnote{We note that the logarithm does not change the optimization problem, but it is typically used in practice.}
\begin{equation}
    x^{*} = \arg \max \ln p(y_{data}|x) ,
\end{equation}
where $p(y_{data}|x)$ is the likelihood function. Sometimes, it is more advantageous to calculate the posterior:
\begin{equation}
    \ln p(x|y_{data}) = \ln p(y_{data}|x) + \ln p(x) - \ln p(y_{data}),
\end{equation}
where $p(x)$ denotes the prior over $x$, and $p(y_{data})$ is the marginal likelihood. The posterior $p(x|y_{data})$ could be further used in Bayesian inference.

In many practical applications, the likelihood function is unknown, but it is possible to obtain (approximate) samples from $p(y|x)$ through a simulator. Such a problem is referred to as \textit{likelihood-free inference} \citep{cranmer2020frontier} or a black-box optimization problem \citep{audet2017derivative}. If the problem is about finding the posterior distribution over $x$ while only a simulator is available, then it is considered as an \textit{Approximate Bayesian Computation} (ABC) problem.

\section{Population-based MCMC}
Typically, a likelihood-free inference problem or an ABC problem is solved through sampling. One of the most well-known sampling methods is the Metropolis-Hastings algorithm \citep{metropolis1949monte}, where the samples are generated from an ergodic Markov chain, and the target density is estimated via Monte Carlo sampling. In order to speed up computations, it is proposed to run multiple chains in parallel rather than sampling from a single chain. This approach is known as \textit{population-based MCMC methods} \citep{iba2001population}. A population-based MCMC method operates over a joint state space with the following distribution:
\begin{equation}
    p(x_{1}, \ldots, x_{C}) = \prod_{c \in \mathcal{C}} p_{c}(x_{c})
\end{equation}
where $\mathcal{C}$ denotes the population of chains, and at least one of $p_{c}(x_{c})$ is equivalent to the original distribution we want to sample from (e.g., the posterior distribution $p(x|y_{data})$).

Given a population of chains, a question of interest is what is the best proposal distribution for an efficient sampling convergence. One approach is \textit{parallel tempering}. It introduces an additional temperature parameter, and initializes each chain at a different temperature \citep{hukushima1996math, liang2000evolutionary}. However, the performance of the algorithm highly depends on an appropriate cooling schedule rather than a \textit{smart} interaction between the chains. A different approach proposed by \cite{strens2002markov}, relies on a suitable proposal that is able to adapt the shape of the population at a single temperature. We further expand on this idea by formulating population-based proposal distributions that are inspired by evolutionary algorithms. 

\subsection{Continuous case}
\cite{ter2006markov} successfully formulates a new proposal called Differential-Evolution Markov Chain (DE-MC) that combines the ideas of Differential Evolution and population-based MCMC. In particular, he redefines the DE-1 equation \citep{storn1997differential} by adding \textit{noise}, $\varepsilon$, to it:
\begin{equation}\label{eq:de_original}
    x_{new} = x_{i} + \gamma(x_{j} - x_{k}) + \varepsilon,
\end{equation}
where $\varepsilon$ is sampled from a Gaussian distribution, $\gamma \in \mathbb{R}_{+}$. The addition of noise is necessary in order to meet the requirements of a Markov chain \citep{ter2006markov}:
\begin{itemize}
    \item \textit{Reversibility} is met, because the suggested proposal could be inverted to obtain $x_{i}$.
    \item \textit{Aperiodicity} is met, because the Markov Chain follows a random walk.
    \item \textit{Irreducibility} is solved by applying the noise.
\end{itemize}

The results presented by \cite{ter2006markov} indicate an advantage of DE-MC over conventional MCMC with respect to the speed of calculations, convergence and applicability to multimodal distributions. Therefore, proposing DE as an optimal solution for choosing an appropriate scale and orientation for the jumping distribution of a population-based MCMC.

\subsection{Discrete case}
In this paper, we focus on binary variables, because categorical variables could always be transformed to a binary representation. Hence, the most straightforward proposal for binary variables is the independent sampler that utilizes the product of Bernoullis:
\begin{equation}\label{eq:ind_sampl}
    q(x) = \prod_{d}B(\theta_d),
\end{equation}
where $B(\theta_d)$ denotes the Bernoulli distribution with a parameter $\theta_d$. However, the above proposal does not utilize the information available across the population, hence, the performance could be improved by allowing the chains to interact. Exactly this possibility we investigate in the following section.

\section{Our Approach}
\paragraph{Markov kernels} We propose to utilize the ideas outlined by \cite{ter2006markov}, but in a discrete space. For this purpose, we need to relate the DE-1 equation to logical operators, as now the vector $x$ is represented by a string of bits, $\mathcal{X} = \{0,1\}^{D}$, and properly define noise. Following \citep{strens2003evolutionary}, we propose to use the \textit{xor} operator between two bits $b_1$ and $b_2$:
\begin{equation}
    b_1 \otimes b_2 = \left\{\begin{matrix}
        1, & b_{1} \neq b_2\\ 
        0, & b_{1} = b_2
    \end{matrix}\right.
\end{equation}
instead of the subtraction in (\ref{eq:de_original}). Next, we define a set of all possible differences between two chains $x_i$ and $x_j$, namely, $\Delta = \{\delta_k: \forall_{x_i, x_j \in \mathcal{C}}\ \delta_k = x_i \otimes x_j\}$.\footnote{A similar construction could be done for the continuous case.} We can construct a distribution over $\delta_k$ as a uniform distribution:
\begin{equation}
    q(\delta|\mathcal{C}) = \frac{1}{|\Delta|}\sum_{\delta_{k}\in \Delta} \mathds{I} \Big{[} \delta_{k} = \delta \Big{]}.
\end{equation}
Now, we can formulate a binary equivalence of the DE-1 equation by adding difference drawn from the $q(\delta|\mathcal{C})$:
\begin{equation}
    x_{new} = x_{i} \otimes \delta_k.
    \label{eq:xor}
\end{equation}
Hoever, the proposal defined in (\ref{eq:xor}) is not a valid Markov kernel as it is shown in the following Proposition.
\begin{proposition}\label{prop:xor}
The proposal defined in (\ref{eq:xor}) fulfills \textit{reversibility} and \textit{aperiodicity}, but it does not meet the \textit{irreducibility} requirement.
\end{proposition}
\begin{proof}
\textit{Reversibility} is met, as $x_{i}$ can be re-obtained by applying the difference to the left side of (\ref{eq:xor}). \textit{Aperiodicity} is met because the general set-up of the Markov chain is kept unchanged (it resembles a random walk). However, the operation in (\ref{eq:xor}) is deterministic, thus, it violates the \textit{irreducibility} assumption.
\end{proof}

The missing property of (\ref{eq:xor}) could be fixed by including the following \textit{mutation} (\textit{mut}) operation:
\begin{equation}
    x_{l} = \left\{\begin{matrix}
    1-x_{l} & if \: p_{flip} \geq  u \\ 
    x_{l} & otherwise
    \end{matrix}\right.
    \label{eq:mut}
\end{equation}
where $p_{flip} \in (0, 1)$ corresponds to a probability of flipping a bit, and $U(0,1)$ denotes the uniform distribution. Then, the following proposal could be formulated \citep{strens2003evolutionary} as in Proposition \ref{prop:mut_xor}.
\begin{proposition}\label{prop:mut_xor}
The proposal defined as a mixture $q_{mut+xor}(x|\mathcal{C}) = \pi q_{mut}(x|\mathcal{C}) + (1-\pi) q_{xor}(x|\mathcal{C})$, where $\pi \in (0, 1)$, $q_{mut}(x|\mathcal{C})$ is defined by (\ref{eq:mut}), and $q_{xor}(x|\mathcal{C})$ is defined by (\ref{eq:xor}), is a proper Markov kernel. 
\end{proposition}
\begin{proof}
\textit{Reversibility} and \textit{aperiodicity} were shown in Proposition \ref{prop:xor}. The \textit{irreducibility} is met, because the \textit{mut} proposal assures that there is a positive transition probability across the entire search space.
\end{proof}

However, we notice that there are two potential issues with the mixture proposal \textit{mut+xor}. First, it introduces another hyperparameter, $\pi$, that needs to be determined. Second, for improperly chosen $\pi$, the converngece speed could be decreased if the \textit{mut} proposal is used too often or too rarely.

In order to overcome these issues, we propose to apply the \textit{mut} operation in (\ref{eq:mut}) directly to $\delta_k$, in a similar manner how the Gaussian noise is added to $\gamma (x_i - x_j)$ in the proposition of \cite{ter2006markov}. As a result, we obtain the following proposal:
\begin{equation}
    x_{new} = x_{i} \otimes (\textit{mut}(\delta_k)).
    \label{eq:de-mc-d}
\end{equation}
Importantly, this proposal fulfills all requirements for the Markov kernel.
\begin{proposition}\label{prop:de-mc-d}
The proposal defined in (\ref{eq:de-mc-d}) is a valid Markov kernel.
\end{proposition}
\begin{proof}
\textit{Reversibility} and \textit{aperiodicity} are met in the same manner how it is shown in Proposition \ref{prop:xor}. Adding the mutation operation directly to $\delta_k$ allows to obtain all possible states in the discrete space, thus, the \textit{irreducibility} requirement is met.
\end{proof}
We refer to this new Markov kernel for discrete random variables as \textit{Discrete Differential Evolution Markov Chain} (\textit{dde-mc}). 

\paragraph{Population-MCMC-ABC} Since we have formulated a proposal distribution that utilizes a population of chains, we propose to use a population-based MCMC algorithm for the discrete ABC problems. The core of the MCMC-ABC algorithm is to use a proxy of the likelihood-function defined as an $\epsilon$-ball from the observed data, i.e., $\| y - y_{data} \| \leq \epsilon$, where $\epsilon > 0$, and $\| \cdot \|$ is a chosen metric. The convergence speed and the acceptance rate highly depend on the value of $\epsilon$ \citep{barber2015rate, faisal2013new, ratmann2007using}. In this paper, we consider two approaches to determine the $\epsilon$ value: (i) by setting a fixed value, and (ii) by sampling $\epsilon \sim Exp(\tau)$ \citep{bortot2007inference}. See the appendix for details.

A single step of the Population-MCMC-ABC algorithm is presented in Algorithm \ref{alg:MCMC-ABC}. Notice that in line 4, we take advantage of the symmetricity of all proposal. Moreover, in the procedure, we skip an outer loop over all chains for clarity. Without loosing the generality, we assume a simulator to be a probabilistic program denoted by $\tilde{p}(y|x)$.

\begin{algorithm}
	\caption{Population-MCMC-ABC} 
	\begin{algorithmic}[1]
	    \State $x' \sim q(x'|\mathcal{C})$\Comment{Either (\ref{eq:ind_sampl}), \textit{mut+xor} or \textit{dde-mc}.}
	    \State Simulate $y \sim \tilde{p}(y|x')$.
	    \If{$\left \| y - y_{data} \right \| \leq \epsilon$}
	        \State $\alpha = \min\{1, \frac{p(x')}{p(x)}\}$
	        \State $u \sim U(0,1)$
	        \If{$u \leq \alpha$}
	            \State $x = x'$
	       \EndIf
	   \EndIf
	   \State \textbf{return} $x$
	\end{algorithmic}
	\label{alg:MCMC-ABC}
\end{algorithm}

\section{Experiments}
In order to verify our proposed approach, we use four test-beds: 
\begin{enumerate}
    \item \textit{QMR-DT Network (likelihood-based case): } First, we validate the novel proposal, \textit{dde-mc}, on a problem when the likelihood is known.
    \item \textit{QMR-DT Network (likelihood-free case): } Second, we verify the performance of the presented proposal by modifying the first test-bed as a likelihood-free problem. 
    \item \textit{Binarized Neural Network Learning: } Third, we investigate the performance of the proposed approach on a high-dimensional problem, namely, learning binary neural networks.
    \item \textit{Neural Architecture Search: } lastly, we consider a problem of Neural Network Architecture Search (NAS). 
\end{enumerate}

The code of the methods and all experiments is available under the following link: \url{https://github.com/IlzeAmandaA/ABCdiscrete}

\subsection{A likelihood-based QMR-DT Network}
\paragraph{Implementation Details} The overall set-up is designed as described by \cite{strens2003evolutionary}, i.e., we consider a QMR-DT Network model. The architecture of the network can be described as a two-level or bipartite graphical model, where the top level of the graph contains nodes for the \textit{diseases}, and the bottom level contains nodes for the \textit{findings} \citep{jaakkola1999variational}. The following density model captures the relations between the diseases ($x$) and findings ($y$):
\begin{equation}
    p(y_{i}=1|x) = 1 - (1-q_{i0})\prod_{l}(1-q_{il})^{x_{l}}
    \label{eq:qmr}
\end{equation}
where $y_{i}$ is an individual bit of string $y$, $q_{i0}$ is the corresponding \textit{leak probability}, i.e., the probability that the finding is caused by means other than the diseases included in the QMR-DT model \citep{jaakkola1999variational}. $q_{il}$ is the association probability between disease $l$ and finding $i$, i.e, the probability that the disease $l$ alone could cause the finding $i$ to have a positive outcome. For a complete inference, the prior $p(b)$ is specified. We follow the assumption taken by \cite{strens2003evolutionary} that the diseases are independent:
\begin{equation}
    p(x) = \prod_{l}p_{l}^{x_{l}}(1-p_{l})^{(1-x_{l})}
\end{equation}
where $p_{l}$ is the prior probability for disease $l$. 

We compare the performance of the \textit{dde-mc} kernel to the \textit{mut} proposal, the \textit{mut-xor} proposal, the \textit{mut+crx} proposal (see \citep{strens2003evolutionary} for details), and the independent sampler (\textit{ind-samp}) as in (\ref{eq:ind_sampl}) with $\theta_d = 0.5$. The use of an independent-sampler proposal allows us to verify whether the presented proposals are useful to efficiently sample in discrete spaces. 
Therefore, we expect the DE-inspired proposals to outperform the \textit{ind-samp}, and the \textit{dde-mc} to perform similarly, if not surpass, \textit{mut+xor}. Out of the possible parameter settings we investigate the following population sizes $C = \left \{8,12,24,40,60 \right \}$, as well as \textit{bit-flipping} probabilities $p_{flip}$ = $\left \{0.1, 0.05, 0.01, 0.005 \right \}$. All experiments are run for 10,000 iterations with 80 cross-evaluations initialized with different underlying parameter settings. 

In this experiment, we used the error that is defined as the average Hamming distance between the real values of $x$ and the most probable values found by the Population-MCMC with different proposals. The number of diseases was set to $m=20$, and the number of findings was $n=80$.

\subsubsection{Results \& Discussion}

\begin{figure}[H]
    \centering
    \includegraphics[width=1.0\columnwidth]{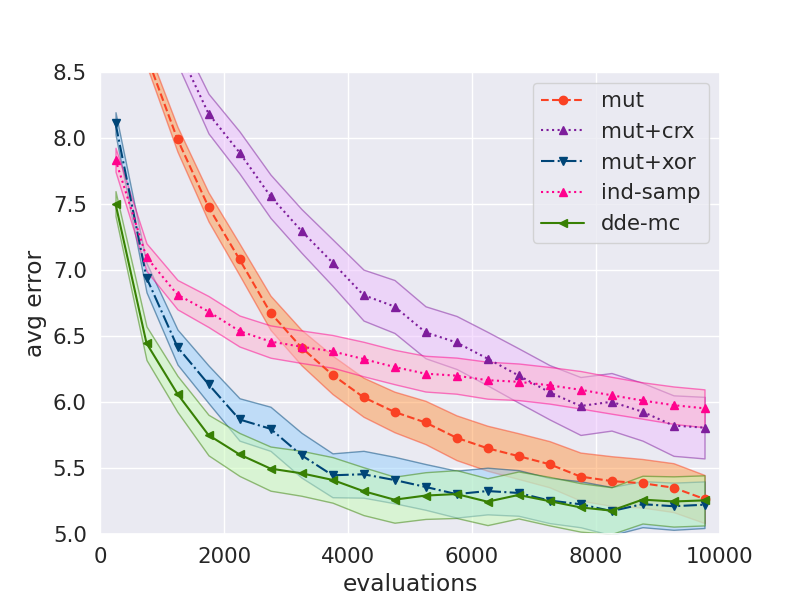}
    \vskip -2mm
    \caption{A comparison of the considered proposals using the population average error (the shaded area corresponds to the standard error).}
    \label{fig:compari}
\end{figure}

 DE inspired proposals, \textit{dde-mc} and \textit{mut+xor}, are superior to kernels stemming from genetic algorithms or random search, i.e., \textit{mut+crx}, \textit{mut} and \textit{ind-samp} (see Figure \ref{fig:compari}). In particular, \textit{dde-mc} converged the fastest (see the first 4,000 evaluations in Figure \ref{fig:compari}), suggesting that an update via a single operator rather than a mixture is most effective. As expected, \textit{ind-samp} requires many evaluations to obtain a reasonable performance.
 
\subsection{A Likelihood-free QMR-DT Network}
\paragraph{Implementation Details} In this test-bed, the QMR-DT network is redefined as a simulator model, i.e., the likelihood is assumed to be intractable. The Hamming distance is selected as the distance metric, but due to its equivocal nature for high-dimensional data, the dimensionality of the problem is reduced. In particular, the number of \textit{diseases} and observations (i.e., \textit{findings}) are decreased to 10 and 20, respectively, while the probabilities of the network are sampled from a Beta distribution, $Beta(0.15,0.15)$. The resulting network is more deterministic as the underlying density distributions are more peaked, thus, the stochasticity of the simulator is reduced. Multiple tolerance values are investigated to find the optimal settings, $\epsilon = \left \{0.5, 0.8, 1., 1.2, 1.5, 2. \right \}$, respectively. The minimal value is chosen to be 0.5 due to variability across the observed data $y_{data}$. Additionally, we checked sampling $\epsilon$ from the exponential distribution. All experiments are cross-evaluated 80 times, and each experiment is initialized with different underlying parameter settings.

\subsubsection{Results \& Discussion}

\begin{figure}[H]
    \centering
    \includegraphics[width=1.0\columnwidth]{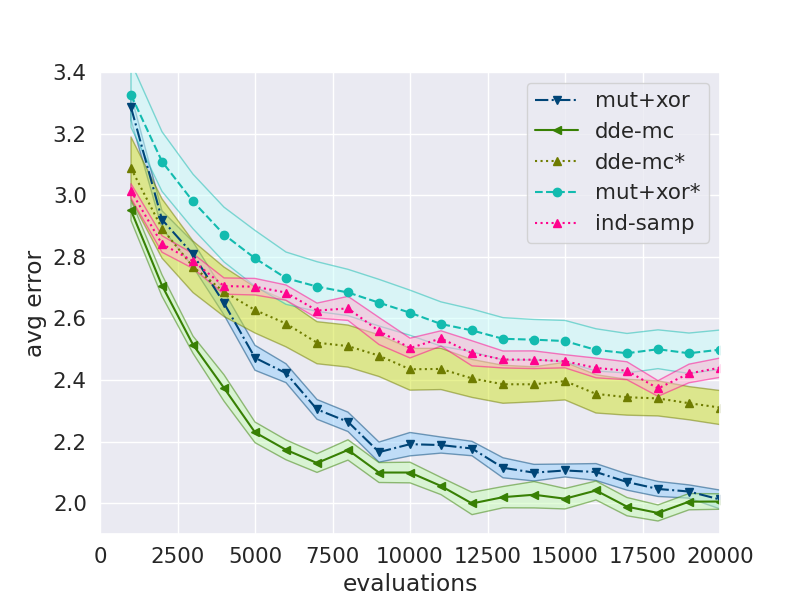}
    \vskip -2mm
    \caption{A comparison of the considered proposals using the population error for exponentially adjusted $\epsilon$ and the fixed $\epsilon$ (\textit{indicated by *}). The shaded area corresponds to the standard error.}
    \label{fig:ep}
\end{figure}

First, for the fixed value of $\epsilon$, we notice that \textit{dde-mc} converged faster and to a better (local) optimum than \textit{mut+xor}. However, this effect could be explained by a lower dimensionality of the problem compared to the first experiment. Second, utilizing the exponential distribution had a profound positive effect on the convergence rate of both \textit{dde-mc} and \textit{mut+xor} (Figure \ref{fig:ep}). This confirmed the expectation that an adjustable $\epsilon$ has a better balance between exploration and exploitation. In particular, $\epsilon \sim Exp(2)$ brought the best results with \textit{dde-mc} converging the fastest, followed by \textit{mut+xor}, and \textit{ind-samp}. This comes in line with the corresponding acceptance rates for the first 10,000 iterations (table \ref{tab:acept_ratio}), i.e., the use of a more \textit{smart} proposal allows to increase the acceptance probability, as the search space is investigated more efficiently. 

\vskip -4mm
\begin{table}[H]
\small
\caption{Probability of the acceptance $\alpha$.}\label{tab:acept_ratio}
\begin{center}
    \begin{tabular}{cc}
     \textbf{ $proposal$} &\textbf{ mean (std) } \\
      \hline \\
      dde-mc  & 24.47 (1.66) \\
      mut+xor & 25.81 (1.38) \\ 
      ind-samp & 13.14 (0.33) \\
    \end{tabular}
\end{center}
\end{table}
\vspace{-0.5cm}

\begin{figure}[H]
    \centering
    \includegraphics[width=1.0\columnwidth]{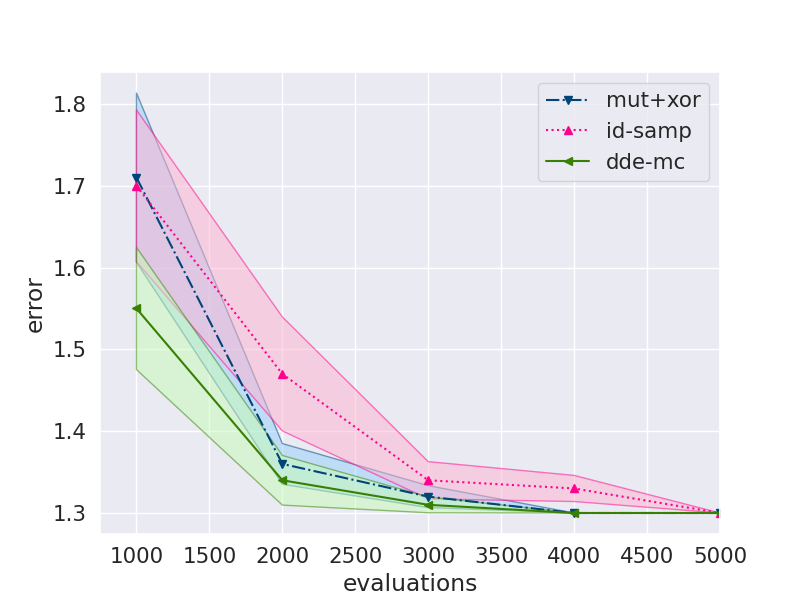}
    \vskip -2mm
    \caption{A comparison of the considered proposal using the minimum error (i.e., the lowest error found by the population) on QMR-DT with adjusted $\epsilon$.}
    \label{fig:qmr_miner}
\end{figure}

Furthermore, the final error obtained by the \textit{likelihood-free} inference approach is comparable with the results reported for the \textit{likelihood-based} approach (Figure \ref{fig:ep} and Figure \ref{fig:compari}). This is a positive outcome as any approximation of the likelihood will always be inferior to an exact solution. In particular, the final error obtained by the \textit{dde-mc} proposal is lower, however, this is accounted by the reduced dimensionality of the problem. Interestingly, despite approximating the likelihood, the computational time  has only increased twice, while the best performing chain is already identified after 4000 evaluations (Figure \ref{fig:qmr_miner}). 

\begin{figure}[H]
    \centering
    \includegraphics[width=1.0\columnwidth]{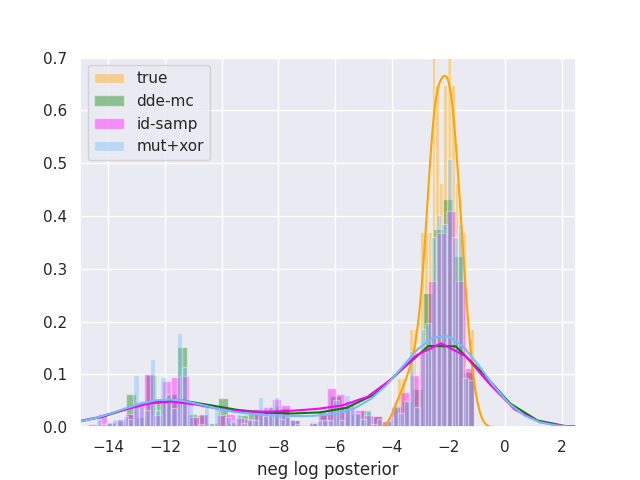}
    \vskip -2mm
    \caption{Approximate posterior distribution}
    \label{fig:dist}
\end{figure}

Lastly, the obtained results were validated by comparing the true posterior distribution of the underlying true values, $x_{true}$, to the final obtained values $x$ by the Population-MCMC-ABC algorithm. In Figure \ref{fig:dist} the negative logarithm of the posterior distribution is plotted. The main conclusion is that all proposals converge towards the approximate posterior, yet the obtained distributions are more dispersed. 

\subsection{Binary Neural Networks}

\paragraph{Implementation Details} In the following experiment, we aim at evaluating our approach on a high-dimensional optimization problem. We train a binary neural network (BinNN) with a single fully-connected hidden layer on the image dataset of ten handwritten digits (MNIST \cite{lecun1998gradient}). We use 20 hidden units and the image is resized from 28px $\times$ 28px to 14px $\times$ 14px. Furthermore, the image is converted to polar values of +1 or -1,  while the the network is created in accordance to \cite{courbariaux2016binarized}, where the weights and activations of the network are binary, meaning that they are constrained to +1 or -1 as well. We simplify the problem to a binary classification by only selecting 2 digits from the dataset. As a result, the total number of weights equals 3940. We use the $tanh$ activation function for the hidden units and the sigmoid activation function for the outputs. Consequently, the distance metric becomes the classification error:
\begin{equation}
    \left \| y_{data} - y  \right \| = 1 - \frac{1}{N} \sum_{n=1}^{N} \mathds{I}\left [ y_{n} = y_{n}(x) \right ] ,
    \label{eq:nn_error}
\end{equation}
where $N$ denotes the number of images, $\mathds{I}[\cdot]$ is an indicator function, $y_{n}$ is the true label for the \textit{n}-th image, and $y_{n}(x)$ is the \textit{n}-th label predicted by the binary neural net with weights $x$.

For the Metropolis acceptance rule we define a Boltzmann distribution over the prior distribution of the weights $x$ inspired by the work of \cite{tomczak2015probabilistic}:
\begin{equation}
    p(x) = \frac{h(x)}{\sum_{i} h(x_i)} ,
\end{equation}
where $h(x) = exp(-\frac{1}{D}\sum_{i \in D} x_{i})$. As a result, the prior distribution acts as a regularization term as it favors parameter settings with fewer \textit{active} weights. The distribution is independent of the data $y$, thus, the partition function $\sum_{i} h(x_{i})$ cancels out in the computation of the metropolis ratio:
\begin{equation}
    \alpha = \frac{p(x')}{p(x)} = \frac{h(x')}{h(x)} .
\end{equation}

The original dataset consists of 60,000 training examples and 10,000 test examples. For our experiment we select the digits 0 and 1, hence, the dataset size is reduced to 12,665 training and 2,115 test examples. Different tolerance values are investigated to obtain the best convergence, ranging from 0.03 to 0.2, and each experiment is ran for at least 200,000 iterations. All experiments are cross-evaluated 5 times. Lastly, we evaluate the performance by computing both the minimum test error obtained by the final population, as well as the test error obtained by using a Bayesian approach, i.e., we compute the true predictive distribution via majority voting by utilizing an ensemble of models. In particular, we select the five last updated populations, resulting in 5x24x5=600 models per run, and we repeat this with different seeds 10 times.

Because the classification error function in (\ref{eq:nn_error}) is non-differentiable, the problem could be treated as a black-box objective. However, we want to emphasize that we do not propose our method as an alternative to gradient-based learning methods. In principle, any gradient-based approach will be superior to a derivative-free method as what a derivative-free method tries to achieve is to implicitly approximate the gradient \citep{audet2017derivative}. Therefore, the purpose of the presented experiment is not to showcase a state-of-the-art classification accuracy, as that already has been done with gradient-based approaches for BinNN \citep{courbariaux2016binarized}, but rather showcase the Population-MCMC-ABC applicability to a high-dimensional optimization problem. 

\subsubsection{Results \& Discussion}

\begin{figure}[H]
    \centering
    \includegraphics[width=1.\columnwidth]{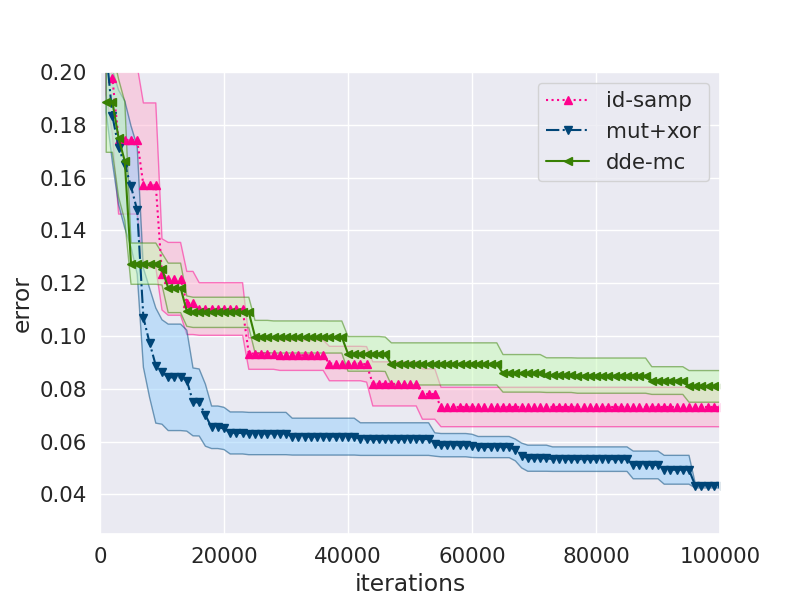}
    \vskip -2mm
    \caption{A comparison of the considered proposals using the minimum training error on MNIST.}
    \label{fig:abc_bnn_min}
\end{figure}

For the high-dimensional data problem, the \textit{mut+xor} proposal converged the fastest towards the optimal solution in the search space (Figure \ref{fig:abc_bnn_min}). In particular, the minimum error on the training set was already found after 100,000 iterations, and a tolerance threshold of 0.05 had the best trade off between Markov chain error and the likelihood approximation bias. 

With respect to the error within the entire population, \textit{dde-mc} converged the fastest, although, its performance is on par with \textit{ind-samp}, while \textit{mut+xor} performs the worst. In general, the drop in performance with respect to the convergence rate of the entire population could be explained by the high dimensionality of the problem, i.e., the higher the dimensionality, the more time is needed for every chain to explore the search space. This observation is confirmed by computing the test error via utilizing all the population members in a majority-voting setting. In particular, the test error based on the ensemble approach is alike across all three proposals, yet the minimum error (i.e., for a single best model) is better for \textit{dde-mc} and \textit{mut+xor} compared to \textit{ind-samp} (Table \ref{tab:bnn}). This result suggests that there seems to be an added advantage of utilizing DE-inspired proposals in faster convergence towards a local optimal solution.

\begin{figure}[H]
    \centering
    \includegraphics[width=1.0\columnwidth]{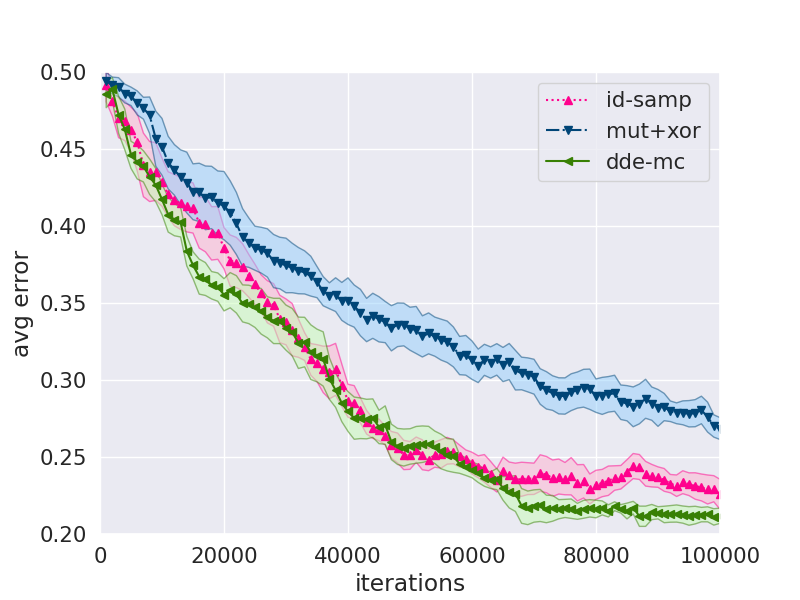}
    \vskip -2mm
    \caption{A comparison of the considered proposals using the avg. training error on MNIST.}
    \label{fig:abc_bnn_pop}
\end{figure}

\begin{table}[H]
\small
\caption{Test error of BinNN on MNIST.}\label{tab:bnn}
\begin{center}
    \begin{tabular}{ccc}
     \textbf{proposal} & \multicolumn{2}{c}{\textbf{error} (ste)} \\
     & \textit{single best} & \textit{ensemble}\\
      \hline \\
       dde-mc  & 0.045 (0.002) & 0.013 (0.001)\\
       mut+xor & 0.046 (0.002) & 0.014 (0.002)\\ 
       ind-samp & 0.051 (0.002) &  0.012 (0.001) \\
    \end{tabular}
\end{center}
\end{table}

\subsection{Neural Architecture Search}
\paragraph{Implementation Details} In the last experiment, we aim at investigating whether the proposed approach is applicable for efficient neural architecture search. In particular, we make use of the NAS-Bench-101 dataset, the first public architecture dataset for NAS research \citep{ying2019bench}. The dataset is represented as a table which maps neural architectures to their training and evaluations metrics, and, as such, it represents an efficient solution for querying different neural topologies. Each topology is captured by a directed acyclic graph represented by an adjacency matrix. The number of vertices is set to 7, while the maximum amount of edges is 9. Apart from these restrictions, we limit the search space by constricting the possible operations for each vertex. Consequently, the simulator is captured by querying the dataset, while the distance metric now is simply the validation error. The prior distribution is kept the same as for the previous experiment. 

Every experiment is ran for at least 120,000 iterations, with 5 cross-evaluations. To find the optimal performance, the following tolerance threshold values are investigated $\epsilon=\left \{ 0.01, 0.1, 0.2, 0.3 \right \}$. As we are approaching the problem as an optimization task, the aim is to find a chain with the lowest test error, rather than covering the entire distribution. Therefore, to evaluate the performance we will plot the minimum error obtained through the training process, as well as the lowest test error obtained by the final population. 

\subsubsection{Results \& Discussion}

\begin{figure}[H]
    \centering
    \includegraphics[width=1.0\columnwidth]{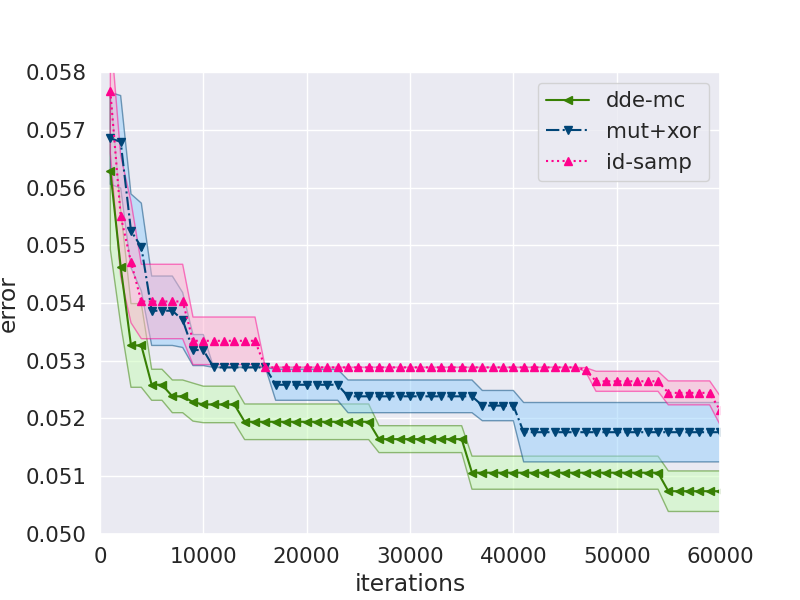}
    \vskip -2mm
    \caption{A comparison of the considered proposals using the minimum training error on NAS-Bench-101.}
    \label{fig:nac_training}
\end{figure}

\textit{Dde-mc} identified the best solution the fastest with $\epsilon$ set to $\epsilon \sim Exp(0.2)$ (fig \ref{fig:nac_training}). The corresponding test error is reported in table \ref{tab:nas}, and it follows the same pattern, namely, \textit{dde-mc} is superior. Interestingly, here the \textit{mut+xor} proposal performs almost on par with the \textit{ind-samp} proposal for the first $10,000$ iterations, and then both methods converge to almost the same result. Our proposed Markov kernel, obtains again not only the best result, but also it is the fastest.

\begin{table}[H]
\small
\caption{Test error on NAS-Bench-101.}\label{tab:nas}
\begin{center}
    \begin{tabular}{cc}
     \textbf{proposal} &\textbf{error} (ste) \\
      \hline \\
       dde-mc  & 0.058 (0.001) \\
       mut+xor & 0.060 ($<0.001$) \\ 
       ind-samp & 0.062 ($<0.001$) \\
    \end{tabular}
\end{center}
\end{table}

\section{Conclusion}
In this paper, we note that there is a gap in the available methods for likelihood-free inference on discrete problems. We propose to utilize ideas known from evolutionary computing similarly to \citep{ter2006markov}, in order to formulate a new Markov kernel, \textit{dde-mc}, for a population-based MCMC-ABC algorithm. The obtained results suggest that the newly designed proposal is a promising and effective solution for intractable problems in a discrete space.

Furthermore, Markov kernels based on Differential Evolution are also effective to traverse a discrete search space. Nonetheless, great attention has to be paid to the choice of the tolerance threshold for the MCMC-ABC methods. In other words, if the tolerance is set too high, then the performance of the DE-based proposals drops to that of an independent sampler, i.e., the error of the Markov Chain is high. For high-dimensional problems, the proposed kernel seems to be most promising, however, its population error becomes similar to that of \textit{ind-samp}. This is accounted by the fact that for high dimensions it takes more time for the entire population to converge.

In conclusion, we would like to highlight that the present work offers new research directions:
\begin{itemize}
    \item Alternative ABC algorithms like SMC should be further investigated.
    \item In this work, we focused on calculating distances in data space. However, utilizing summary statistics is almost an obvious direction for future work.
    \item As the whole algorithm is based on logical operators, and the input variables are also binary, the algorithm could be encoded using only bits, thus, saving considerable amounts of memory storage. Consequently, any matrix multiplication could be replaced by a XNOR operation followed by a sum, thus, reducing the computation costs and possibly allowing to implement the algorithm on relatively simple devices. Therefore, a natural consequence of this work would be a direct hardware implementation of the proposed methods.
    
\end{itemize}

\bibliography{bib}

\section*{Appendix}
\subsection*{$\epsilon$ determination}
The choice of $\epsilon$ defines which data points are going to be accepted, as such it implicitly models the likelihood. Setting the value too high will result in a biased estimate, however it will improve the performance of Monte Carlo as more samples are utilized per unit time. Hence, as \cite{lintusaari2017fundamentals}, already has stated: "the goal is to find a good balance between the bias and the Monte Carlo error". 

\paragraph{Fixed $\epsilon$} The first group of tolerance selection methods are all based on a fixed $\epsilon$ value. The possible approaches are summarized as follows:
\begin{itemize}
    \item \textit{determine a desirable acceptance ratio:} For example define a proportion, 1\%,  of the simulated samples that should be accepted (\cite{beaumont2002approximate}).
    \item \textit{re-use generated samples:} Determine the optimal cutoff value by a leave-one-out cross validation approach of the underlying parameters of the generated simulations. In particular, minimize the root mean squared error (RMSE) for the validation parameter values \citep {faisal2013new}.
    \item \textit{use a pilot run to tune:} Based on rates of convergence \cite{barber2015rate}, defines fixed alterations to the initial tolerance value in order to either increase the number of accepted samples, reduce the mean-squared error, or increase the (expected) running time. 
    \item \textit{set $\epsilon$ to be proportional to $N_{s}^{-1/(d+5)}$:} where d is the number of dimensions (for a complete overview see \citep{lintusaari2017fundamentals}). 
\end{itemize}
Nonetheless, setting $\epsilon$ to a fixed value hinders the convergence as it clearly is a sub-optimal approach due to its static nature. Ideally, we want to promote exploration in the beginning of the algorithm and, subsequently, move towards exploitation. Hence, alluding to the second group of tolerance selection methods: adaptive $\epsilon$. 

\paragraph{Adaptive $\epsilon$}
In general, the research on adaptive tolerance methods for MCMC-ABC is very limited as traditionally adaptive tolerance is seen as part of SMC-ABC. In the current literature two adaptive tolerance methods for MCMC-ABC are mentioned:
\begin{itemize}
    \item \textit{an exponential cooling scheme:} \cite{ratmann2007using} suggest using an exponential temperature scheme combined with a cooling scheme for the covariance matrix $\sum_{t}$.
    \item \textit{sample from exponential distribution:} Similarly, \cite{bortot2007inference} assume a pseudo-prior for $\epsilon$ : $\pi(\epsilon)$, where $\pi(\epsilon)\sim Exp(\tau)$ and $\tau = 1/10$. Thus allowing to occasionally generate larger tolerance values to adjust mixing. 
\end{itemize}

In order to establish a clear baseline for MCMC-ABC in a discrete space we decided to implement both fixed and adaptive $\epsilon$. Such an approach allows us to evaluate what is the effect of an adaptive $\epsilon$ in comparison to a fixed $\epsilon$ in a discrete space, as well as to compare how well our observations come in line with the observations drawn in a continuous space. 

\end{document}